\documentclass[letterpaper]{article} % DO NOT CHANGE THIS
\usepackage[submission]{aaai24}  % DO NOT CHANGE THIS
\usepackage{booktabs}
\usepackage{times}  % DO NOT CHANGE THIS
\usepackage{helvet}  % DO NOT CHANGE THIS
\usepackage{courier}  % DO NOT CHANGE THIS
\usepackage[hyphens]{url}  % DO NOT CHANGE THIS
\usepackage{graphicx} % DO NOT CHANGE THIS
\usepackage{natbib}  % DO NOT CHANGE THIS AND DO NOT ADD ANY OPTIONS TO IT
\usepackage{caption} % DO NOT CHANGE THIS AND DO NOT ADD ANY OPTIONS TO IT
\usepackage{algorithm}
\usepackage{algorithmic}
\usepackage{mathptmx}
\usepackage{graphicx}
\usepackage{amsmath}
\usepackage{amssymb}
\usepackage{booktabs}
\usepackage{multirow}
\usepackage{colortbl}
\usepackage{newfloat}
\usepackage{listings}
\DeclareCaptionStyle{ruled}{labelfont=normalfont,labelsep=colon,strut=off} % DO NOT CHANGE THIS
\floatstyle{ruled}
\newfloat{listing}{tb}{lst}{}
\floatname{listing}{Listing}
\usepackage{bibentry}
\begin{document}

\title{Progressive Conservative Adaptation for Evolving Target Domains}

\author
{\textbf{Gangming Zhao}$^{1,2}$,
\textbf{Chaoqi Chen}$^{1}$,
\textbf{Wenhao He}$^{3}$, \\
\textbf{Chengwei Pan}$^{4}$,
\textbf{Chaowei Fang}$^{2}$,
\textbf{Jinpeng Li}$^{3}$, Xilin Chen$^{5}$, and, \textbf{Yizhou Yu}$^{1}$ \\ %Institute of Automation, Chinese Academy of Sciences
\normalsize{$^{1}$The University of Hong Kong}\\
\normalsize{$^{2}$Xidian University }\\
\normalsize{$^{3}$University of Chinese Academy of Sciences}\\
\normalsize{$^{4}$Beihang University}\\
\normalsize{$^{5}$Institute of Computing Technology, Chinese Academy of Sciences}
}

\maketitle

\begin{abstract}
Conventional domain adaptation typically transfers knowledge from a source domain to a stationary target domain. 
However, in many real-world cases, target data usually emerge sequentially and have continuously evolving distributions. Restoring and adapting to such target data results in escalating computational and resource consumption over time. 
Hence, it is vital to devise algorithms to address the evolving domain adaptation (EDA) problem, \emph{i.e.,} adapting models to evolving target domains without access to historic target domains. To achieve this goal, we propose a simple yet effective approach, termed progressive conservative adaptation (PCAda). 
To manage new target data that diverges from previous distributions, we fine-tune the classifier head based on the progressively updated class prototypes.
Moreover, as adjusting to the most recent target domain can interfere with the features learned from previous target domains, we develop a conservative sparse attention mechanism. This mechanism restricts feature adaptation within essential dimensions, thus easing the inference related to historical knowledge.
%Furthermore, since adapting to the most recent target domain hampers the features learned from previous target domains, we devise a conservative sparse attention mechanism, which constrains the feature adaptation within the critical dimensions and relieves the inference to the historic knowledge. 
The proposed PCAda is implemented with a meta-learning framework, which achieves the fast adaptation of the classifier with the help of the progressively updated class prototypes in the inner loop and learns a generalized feature without severely interfering with the historic knowledge via the conservative sparse attention in the outer loop. 
Experiments on Rotated MNIST, Caltran, and Portraits datasets demonstrate the effectiveness of our method.
\end{abstract}

\section{Introduction}

Domain adaptation (DA) leverages the knowledge from a labeled \emph{source} (domain) to an unlabeled \emph{target} (domain) to complete tasks on the target. Most DA algorithms have a stationary assumption that the training and testing samples of the target share the same distribution. Algorithms with such assumption will fail when the target data evolve temporally due to the catastrophic forgetting~\cite{mccloskey1989catastrophic,kirkpatrick2017overcoming}, i.e., when adapting the model for recent domains, the knowledge learned from historic domains may be hampered.

To address this, the evolving domain adaptation (EDA) has been formulated~\cite{liu2020learning}. Figure \ref{EDA} shows EDA paradigm. During training, a number of labeled samples are available as the source and more recent unlabeled samples are available as the target. The online testing data arrive sequentially with varying distributions. The model is not allowed to restore the past testing samples due to the limitations of storage and computation resources. EDA is expected to learn a model capable of adapting to the unseen targets efficiently with very few unlabeled samples during testing.

We divide EDA into two subtasks: 1) how to alleviate the distributional differences between domains to guarantee the performance on the target task (like many DA works, we focus on the image classification), and 2) how to effectively reduce the catastrophic forgetting under limited computing resources.
A typical solution for the first subtask is adversarial training~\cite{ganin2016domain,tzeng2017adversarial,zhao2018adversarial,chen2022reusing}, which learns domain invariant features by playing the minimax game between a feature extractor (viewed as a generator) and a domain discriminator. However, they rely on the domain discriminator to distinguish the discrete domains, which is inapplicable in EDA since the target domain is continuous. On the other hand, the existing EDA method~\cite{liu2020learning} has an unstable domain adaptation process and can not make full use of the information of the evolving domain.
Many methods have been proposed to accomplish the second subtask, such as the memory-based methods ~\cite{rebuffi2017icarl,wu2018memory,zhong2019invariance} or buffer-based methods~\cite{wang2021continual,saha2021saliency,cai2021online}. 
However, these methods are not feasible in the EDA setting and encounter deployment difficulties with limited computing resources. It is noteworthy that ~\cite{liu2020learning} devises a method to solve this problem by using the output of the historical model to regularize the output of the current model. However, the offset of distributions gradually increases as the target domain evolves temporally.

In this paper, we exploit and explore the progressive conservative adaptation (PCAda) algorithm with two core designs. First, to deal with the dynamic/continuous domain problem, we introduce a prototypical vector to each class to capture discriminative features and then design a progressive updating pipeline to adapt the prototypes, i.e., continually updating the prototypical representations with few unlabeled online target samples. The prototypes are applied to adapt the classification head to new target domains. Second, to reduce the interference of the key channels of historical domain features, we propose a conservative sparse attention module to make it focus on updating the most important channels in the current domain. PCAda is implemented with a meta-learning framework. During the meta-training/testing, we alternatively update the classifier with the progressive updating and update the features by the conservative sparse attention. Experimental results indicate that PCAda can effectively learn the evolving domain information and solve the catastrophic forgetting problem.

\begin{figure}[t]
\centering
\includegraphics[scale=0.13]{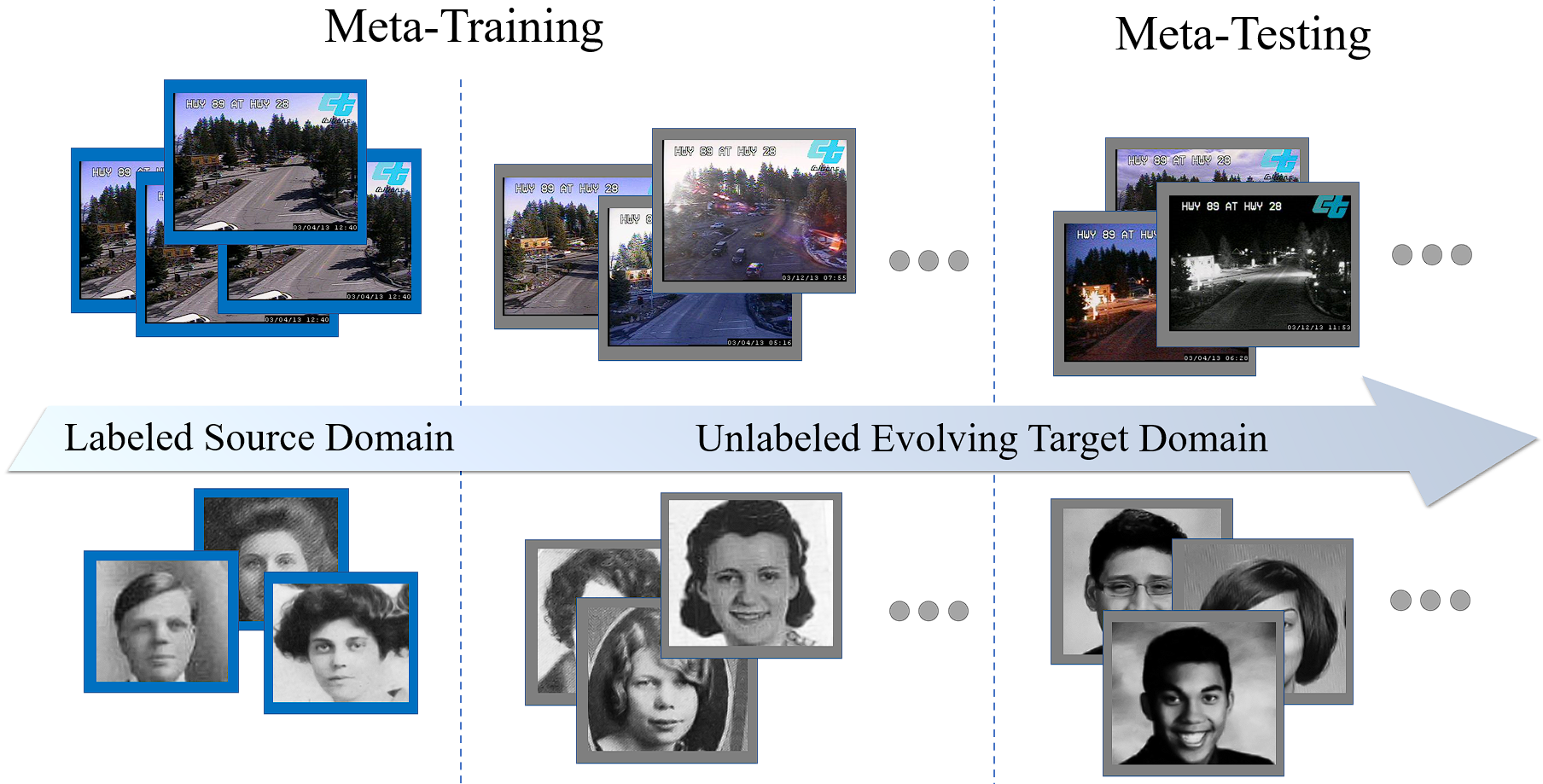}
\caption{\textbf{The Evolving Domain Adaptation (EDA) paradigm.} During meta-training, a small number of labeled source data and a portion of evolving unlabeled target data are accessible; During meta-testing, new target data evolve and the source is inaccessible. We expect the model to perform well on the online target data that comes continuously.}
\label{EDA}
\vspace{-1.5em}
\end{figure}

\section{Related Work}

\textbf{Domain Adaptation.} DA~\cite{pan2010domain,patel2015visual} aims at training a model for a unlabeled target with the help of one or multiple labeled sources. The target usually has evident distribution discrepancy in contrast to the source although they are related. A typical DA algorithm involves two optimization goals: 1) enhancing the accuracy on the source, and 2) reducing the domain discrepancy between the source and target. The first goal can be fulfilled with supervised training strategies, e.g., minimizing the cross-entropy in classification tasks, and the main innovations are made on the second goal. Some works reduce the domain discrepancy with explicit metric. For example, ~\cite{gretton2012kernel,tzeng2014deep,long2015learning} introduced the maximum mean discrepancy (MMD) method to calculate the distance between the source and target and expected them to be as close as possible.
JAN~\cite{long2017deep} proposed a joint distribution difference to directly compare the joint distributions of domains based on MMD. 
Others attempt to reduce the domain discrepancy with implicit metric. For example, the adversarial-based methods~\cite{ganin2016domain,tzeng2017adversarial,zhao2018adversarial,chen2022reusing} calibrate the distribution of the source and target by generating codes that cannot be distinguished by the discriminator, which is trained to classify domains. Although being successful in many applications, these methods are for discrete targets and cannot be applied directly to the EDA problem.

\noindent\textbf{Continuous Domain Adaptation (CDA).} Similar to the EDA, CDA treats each target sample as possibly coming from a different subspace on the manifold of a nonstationary domain. 
Continuous manifold adaptation (CMA)~\cite{hoffman2014continuous} was an early work that considered the adaptation to continuous domains. 
As an extension of CMA, ~\cite{mancini2019adagraph} used the training data collected at some historical points to distinguish classes, whereas the catastrophic forgetting~\cite{mccloskey1989catastrophic,kirkpatrick2017overcoming} remained unsolved.
~\cite{bobu2018adapting,saporta2022multi} pointed out that the incremental adaptation process was prone to catastrophic forgetting, but it did not address the evolutionary features of the target. The CDA methods require the access to both the source and target domains to align the distributions. In this paper, we consider a more complex scenario where the testing is conducted without the access to the source.

\noindent\textbf{Evolving Domain Adaptation.} EDA was formulated in recent years~\cite{liu2020learning}. During the meta-training phase, the model can access the labeled samples from the source and some unlabeled data from the target that evolves over time. During the testing phase, new target data arrive sequentially from the same evolving target distribution. Note that the historical target data and the source data are inaccessible in EDA setting. ~\cite{liu2020learning} presented a meta-adaptation framework called EAML to enable learners to adapt to evolving target domains. 
Based on but quite different from EAML, the proposed PCAda is a more effective solution that can make full use of the inter-domain information with special designs to alleviate the catastrophic forgetting problem. Experimental results demonstrate that our proposed PCAda performs better.

\section{Method}

This paper is targeted at tackling the evolving domain adaptation problem for image classification. Given a labeled source domain image dataset $D_s=\{X^s,Y^s\}$ and an unlabeled target domain image dataset $D_t$, the goal is to train an image classification network that can easily adapt to unknown domains in the future.
Here, $X^s=\{x^s_i\}_{i=1}^{n_s}$ is the set of source domain images with $n_s$ samples, and $Y^s$ denotes the corresponding labels, namely $\{y^s_i\}_{i=1}^{n_s}$. We assume the image distribution evolves continuously with respect to the temporal axis.
The unlabeled target domain dataset can be decomposed into $m$ subdomains with diversified data distributions, namely $D_t= \{X^{t_1},X^{t_2},…,X^{t_m}\}$.
We propose a novel framework to tackle this problem, which can capture the evolving trend of target domains with subtle interference in the feature representations in historic domains. It can also progressively adapt the classifier towards evolving target domains with the help of online class prototypes.

\subsection{Evolving Domain Adaptation in Meta-learning}
Directly applying classical discrete DA methods to evolutionary targets cannot appropriately solve the EDA problem. The reason is twofold. First, discrete DA methods ignore the dynamic characteristic of the target. 
Second, we expect the model to perform well on all the target data since the target data arrive sequentially. We should specify a training strategy to fit the evolution of the target.
The goal of EDA is to learn a model $f_\theta$ parametrized by $\theta$ with continual well performance on the evolving target.
%separately and in succession

Classical theory ~\cite{ben2006analysis,ben2010theory} indicate that to solve the EDA problem with a target trajectory, the target needs to be steadily evolved to facilitate the knowledge transfer. 
To generalize to evolving targets, the EDA method should adapt to random samples of the target trajectory rather than adapting to a single target. Let $d_{H \Delta H}$ denote the divergence between domains, where 
\begin{equation}
    \begin{aligned}
        d_{H \Delta H}(D_s, D_t) =& \sup _{\theta, \theta^{\prime}}\left|\mathbb{E}_{D_s} L\left(f_{\theta}(x), f_{\theta^{\prime}}(x)\right)\right. \\ & -  
         \left.\mathbb{E}_{D_{t}} L\left(f_{\theta}(x), f_{\theta^{\prime}}(x)\right)\right|
    \end{aligned}
\end{equation}

It quantifies the difference between the source and target distributions. $\alpha$ represents the rate of evolution of the target domain $D_t$. 
A fairly small $\alpha$ means that $D_t$ is evenly developed and adjacent target data share knowledge.
We further assume that 
\begin{equation}
    \begin{aligned}
        d_{H \Delta H}(D_{t_1}, D_{t_2}) \leq  \alpha| t_1 - t_2|
    \end{aligned}
\end{equation}
holds with constant $\alpha$ for $t_1, t_2 \ge 0$.
So for any $\theta$ with probability at least $1 - \delta$ over the sampling of the target trajectory $t_{1}, t_{2} \cdots t_{n}$,
% \begin{equation}
\begin{flalign}\label{EDA}
&&
\begin{array}{cc}
\mathbb{E}_{t} \mathbb{E}_{D_{t}} L\left(f_{\theta}(x), y\right) \leq \mathbb{E}_{D_{s}} L\left(f_{\theta}(x), y\right)+ \\
\frac{1}{n} \sum_{i=1}^{n}\left[d_{H \Delta H}\left(D_s, D_{t_i}\right)\right]+\mathbb{E}_{t} \lambda_{t}+O\left(\frac{\alpha}{\delta n}\right)
\end{array}
&&
\end{flalign}
% \end{equation}
where,

\begin{equation}
\lambda_{t}=\min _{\theta}\left[\mathbb{E}_{D_s} L\left(f_{\theta}(x), y\right)+\mathbb{E}_{D_{t}} L\left(f_{\theta}(x), y\right)\right]
\end{equation}
is the adaptability measuring the likelihood of cross-domain learning.
It guarantees the model to learn evolutionary domains without forgetting the source domain.

Considering that the target data can only be read and calculated separately, in succession, how to prevent the forgetting of historical information is a key problem. ~\cite{liu2020learning} used the output of the historical model to regularize the output of the current model, however, this method is unstable, i.e., with the passing of time, the feature offset will gradually increase. We solve this problem from a new perspective by adopting a conservative way. The algorithm focuses on updating the feature channels that play a decisive role in the current domain, and reduces the interference of the key channels in the feature of the historical domain.

Another key issue is how to adjust the classifier quickly. The category feature should change along with the arrival of new data. To implement this function, we designed a progressive prototype updating mechanism for classes that can quickly adapt to changing target domains.

The latest developments in meta-learning~\cite{finn2017model,raghu2019rapid,flennerhag2021bootstrapped} yield transferable features that are able to quickly adapt to subsequent tasks. 
Inspired by this, we consider adopting the meta-learning framework to build the EDA framework. 
Taking model agnostic meta-learning (MAML)~\cite{finn2017model} as an example, it learns transferable features by minimizing errors on the support set in the inner loop and minimizing errors on the query set in the outer loop.
We consider the unlabeled target trajectory from the support set as $D^{spt}_t= \{\hat{X}^{t1},\hat{X}^{t2},…,\hat{X}^{tm}\}$ and the unlabeled target trajectory from the query set as $D^{qry}_t  = \{\widetilde{X}^{t1},\widetilde{X}^{t2},…,\widetilde{X}^{tm}\}$. Note that each $t_i$ in the support set and query set is the same. 

\begin{figure*}[t]
\centering
\includegraphics[height=10cm,width=16cm]{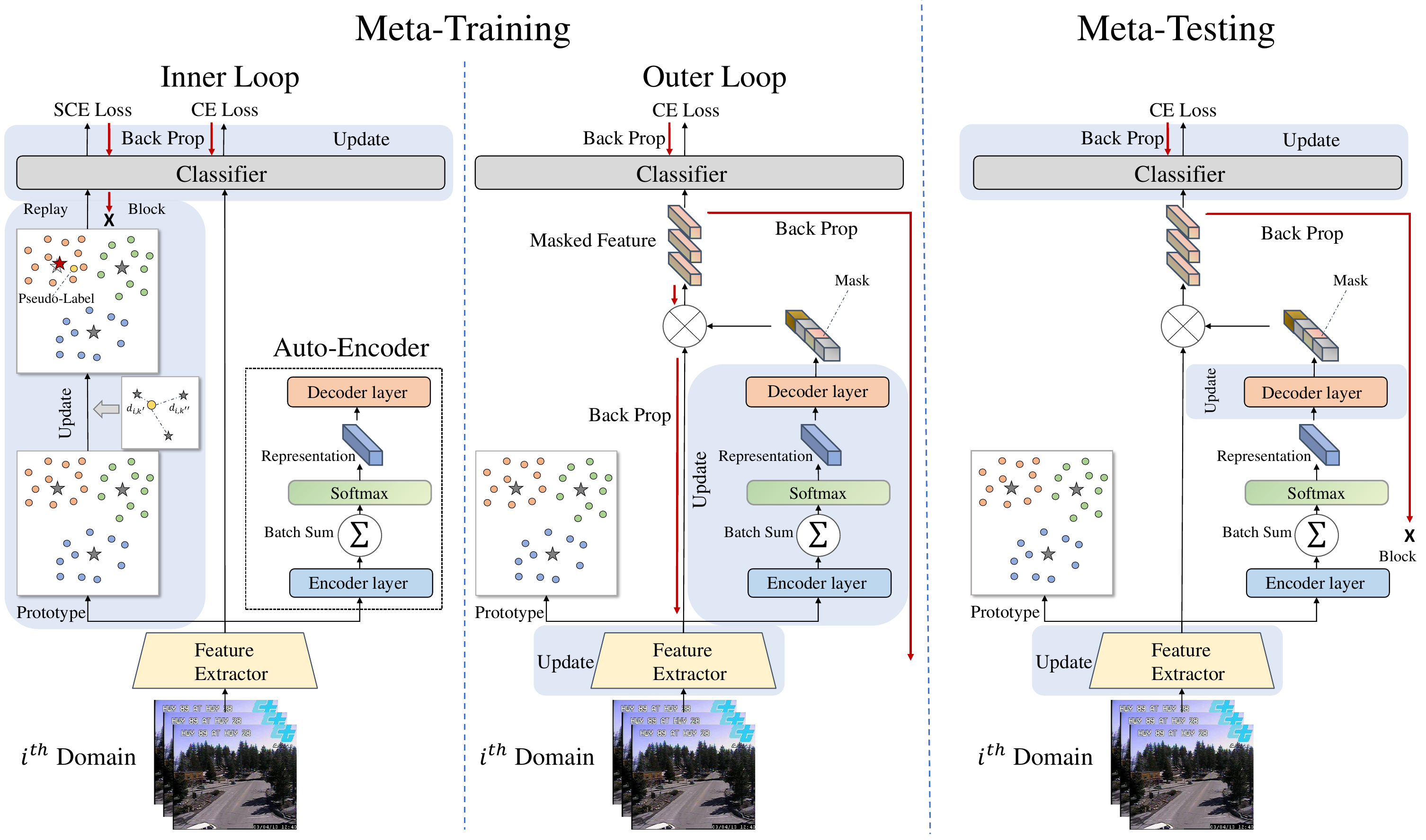}\caption{\textbf{Overview of the PCAda}. In the inner loop, the prototype-based replay updates the classification head; In the outer loop, the sparse attention mechanism helps the feature extractor adapt to the evolving domain. In meta-testing, we fix the encoder and update the classification head, decoder, prototype and feature extractor in the changing target domain.}
\label{structures}
%\vspace{-2em}
\end{figure*}

Before deploying the model to an evolving target domain, we use the source domain data to pre-adapt the model.
Then the model learns knowledge in the inner and the outer loops to better accommodate domains in evolution. The overall structure is shown in Figure ~\ref{structures}.

\raggedbottom

\subsection{Progressive Adaptive Prototype in Inner Loop}
As the number of domains increases, the features of each category change.
Since upcoming data cannot be stored, we propose the adaptive prototype mechanism (APM) to make a fast adaptation to the classification head $f_\theta$ in the inner loop.
The prototype~\cite{snell2017prototypical} corresponding to the labeled data in the source is computed with the following formula:
\setlength{\abovedisplayskip}{3pt}
\setlength{\belowdisplayskip}{3pt}
% \begin{center}
\begin{small}
\begin{flalign}\label{prototype}
&&
\mathbf{c}_{k}=\frac{1}{\left|D_{s}\right|} \sum_{\left({x}^s_{i}, y^s_{i}\right) \in D_{s}} \phi\left({x}^s_{i}\right)
&&
\end{flalign}
\end{small}
% \end{center}
where $\phi$ represents the feature encoder and $c_k$ represents the prototype of class $k$. 
Since the target label is not available, we adopt the active target sample identification~\cite{zhang2019category} to obtain pseudo-labels of unlabeled target data trajectories. 
The pseudo-labels generated by the category anchors are more reliable than the pseudo-labels assigned by the predicted category probabilities. The category anchor is defined as the center of the prototype.

For the unlabeled target datum $x_i^t$, we define the distance between its feature and the center of the $k$-th prototype as
\setlength{\abovedisplayskip}{3pt}
\setlength{\belowdisplayskip}{3pt}
% \begin{center}
\begin{small}
\begin{flalign}\label{prototype_dis}
&&
d_{i,k}=\left\|\phi\left(x_{i}^{t}\right)-\mathbf{c}_{k}\right\|_{2}
&&
\end{flalign}
\end{small}
% \end{center}
where $||.||_2$ represents the 2-norm of the vector. %Then sort $\{d_{k i},k=1,…,K\}$ in ascending order, 
We define the index of the prototype with the closest distance to $x_i^t$ as $k^\prime=\arg\min_{k} d_{i,k}$, and that with the second closest distance to $x_i^t$ as $k^{\prime\prime}=\arg\min_{k,k\neq k^\prime} d_{i,k}$. The pseudo-label of the sample is determined via
\begin{small}
\begin{flalign}
\label{pseudo-label}
&&
\hat{y}_{i}=
\begin{cases}
k^\prime, & d_{i,k^{\prime\prime}}-d_{i,k^\prime} >  \Delta_{d} \\
0, & \textrm{otherwise}
\end{cases}
%k^*,\quad \mbox { if } \;  d^\prime_{i}<d^{\prime\prime}_{i}-\Delta_{d}
&&
\end{flalign}
\end{small}
where $\Delta_d$ is a constant.
%Here $k^*=\arg\min_{k}\{d_{k i},k=1,…,K\}$ .
Then, we use the features of the unlabeled target data to gradually update the prototype of the corresponding category via
\begin{small}
\begin{flalign}\label{update}
&&
\mathbf{c}_{k^\prime}= \mathbf{c}_{k^\prime} + \eta(t) (\hat{y}_{i}>0) [\phi\left(x^t_{i}\right)- \mathbf{c}_{k^\prime}]
&&
\end{flalign}
\end{small}
where $k^\prime$ is the pseudo-label of the active target sample $x_i^t$, and $\eta(t)$ is a deterministic simulated annealing process. If $\eta(t)$ is very small, the pseudo-label prototype will not work, since the model is not adapted to the evolutionary target at the beginning of training. $\eta(t)$ is defined as
\begin{small}
\begin{flalign}\label{tuihuo}
&&
\begin{aligned}
\begin{split}
\eta(t)= \left \{
\begin{array}{ll}
    0  &   t<T_{1}\\
\frac{t-T_{1}}{T_{2}-T_{1}} \eta_{f}  &  T_{1} \leq t<T_{2} \\
    \eta_{f} & t \geq T_{2} \
\end{array}
\right.
\end{split}
\end{aligned}
&&
\end{flalign}
\end{small}
where $\eta_f$ is the balance parameter and $t$ is the training epoch. 
After updating the prototype, we feed them back to the model to force the model to remember useful knowledge learned before. Instead of the standard cross-entropy, we use a more robust symmetric cross-entropy (SCE) loss (\cite{wang2019symmetric}) to further enhance the noise tolerance to stabilize the early training phase.
\begin{small}
\begin{flalign}\label{XX}
&&
L_{in}=\frac{1}{K}\sum_{k=1}^K [a{\ell_{ce}(f_\theta({\mathbf{c}_{k}}),k)}+b{\ell_{ce}(k,f_\theta({\mathbf{c}_{k}}))}]
&&
\end{flalign}
\end{small}
where $a$  and $b$  are the balancing factors and set to 0.2 and 1, respectively. $f_{\theta}$ is the classifier. We use SGD to adapt them to the target data in $D^{spt}_t$:

\setlength{\abovedisplayskip}{3pt}
\setlength{\belowdisplayskip}{3pt}
\begin{small}\label{inner}
\begin{flalign}
&&
    \begin{split}
    \theta_{i+1} \leftarrow \theta_{i}-\alpha_{\text {in }} \nabla_{\theta} \left[L_{1}\left(f_{\theta i}\left(X^{s}\right), Y^{s}\right)+ \right.\\ 
    \left. d\left(f_{\theta i}\left(X^{s}\right), f_{\theta i}\left(\hat{X}^{t i}\right)\right)
    +\eta'(t) L_{in}\right]
    \end{split}
&&
\end{flalign}
\end{small}
where $i$ is the number of iterations for the inner loop and $L_1$ is the cross-entropy loss.
Here, we use the joint maximum mean difference (joint MMD~\cite{long2017deep}) as the domain difference $d$. 
Similar to Equation \ref{update}, the deterministic simulated annealing coefficient $\eta'(t)$ are also used to avoid adding too much noise to the pseudo-labels in the earlier models. $\eta'(t)$ simply replaces $\eta_f$ in Equation \ref{tuihuo}  with $\eta'_f$.
\subsection{Conservative Sparse Attention in Outer Loop}
In the outer loop, we improve the feature extractor $\phi$ with sparse attention. 
Inspired by the meaning of $\lambda_t$ in Equation \ref{EDA}, we need to maintain the model's performance on the evolutionary domain without forgetting the source domain. We approach this task from a new perspective.
% Inspired by the method often adopted in the field of continuous learning (\cite{hinton2006reducing,abati2020conditional}), 
% We consider using sparse mask to multiply the feature representation extracted by the feature extractor for each domain, so as to reduce the ability to forget the old task when learning a new task. 
We consider a relatively conservative way of updating:
based on the sparse attention mechanism (SAM), the model focuses on updating the feature channels corresponding to the current domain and reduces the interference of new data on those channels that play a decisive role in the feature of the historical domain. We argue that applying the method of ~\cite{yang2021generalized} to simply encode the existing domain with an integer is not feasible when considering the setting of incremental evolution. 
%为此我们采取自编码器的架构
Therefore, we adopt the autoencoder structure. ~\cite{hinton2006reducing}. 
For the input of a batch in one target domain, we pass it through the feature extractor and send the output to the encoder layer $A_e$ implemented with a multi-layer perceptron (MLP). We use the batch-sum to get the vector representation. %这里感觉问题还很多
It sums the input data in batch dimension to get a vector representation relative to the batch. 
The goal is to ensure similar (uniform) sparse masks for data in the same target domain.

We use the vector representation through the softmax layer to normalize the vector. The result at this time is the required specific coding representation of the domain, and then it is sent to the decoder layer $A_d$ implemented with MLP to generate a mask for the domain. 
In summary, the output of the embedding layer can be expressed as
\begin{small}
\begin{flalign}\label{embedding}
&&
e_i=A_d(softmax(\sum^B_{j=1}h_{ij}))
&&
\end{flalign}
\end{small}
where $h_{ij}=A_e(x^t_{ij})$ represents the $j$-th sample $x^t_{ij}$ of the $i$-th batch encoded by the encoder $A_e$ and $B$ denotes the batch size.
The sparse attention is a representation close to binary coding,
\begin{small}
\begin{flalign}
&&
A_{\mathrm{i}}=\sigma\left(100 \cdot e_{i}\right)
&&
\end{flalign}
\end{small}
where $\sigma$ is the sigmoid function and the constant 100 is to ensure a near binary output but still differentiable.
Finally, it is multiplied by the output of the previous feature extractor and sent to the classification head.
\begin{small}
\begin{flalign}\label{mask}
&&
\phi'(X^{i})=A_{\mathrm{i}}\cdot\phi(X^{i})
&&
\end{flalign}
\end{small}
The sparse attention mechanism helps model conservative updates to key feature channels in the current domain without forgetting past information. It corresponds to $\lambda_t$ in Equation \ref{EDA}, which makes it possible for the model to adapt to evolving target domains without overwriting critical past knowledge even after training on a fixed source domain.

\textbf{Meta-training}: The model updates the encoder and decoder at the same time in the outer loop so that the encoder can give different intermediate coding representations of different domain data and similar intermediate coding representations of the same domain data. Then model uses the decoder to yield a suitable mask representation.

\textbf{Meta-testing}: In this stage, we block the encoder's gradient back-propagation and only update the decoder and the feature extractor in the outer loop. 
For different domains, the model will only update a portion of the network to prevent  the forgetting.

The EDA method should also learn representations to capture and exploit the evolution of the target domain.
Referring to the method of~\cite{liu2020learning}, we use the upper bound of Equation \ref{EDA} to replace the original loss, and use $\max _{i} d\left(f_{\theta i}\left(\widetilde{X}^{t_{i-1}}\right), f_{\theta i}\left(\widetilde{X}^{t_{i}}\right)\right)$ as the approximate value of $\alpha$ to control the upper bound to minimize the EDA loss on the query set in Equation \ref{EDA}. 
Combined with the proposed attention mechanism, we train the feature extractor $\phi$  and auto-encoder $A$ in the outer loop. Therefore, we consider the following loss:
\begin{small}
\begin{flalign}
&&
\begin{aligned}
L_{\text {out}}=L_{1}\left(f_{\theta i}\left(\widetilde{X}^{s}\right), f_{\theta i}\left(\widetilde{X}^{t_{i}}\right)\right)&+
\\ \frac{1}{n} \sum_{i=1}^{n} d\left(f_{\theta i}\left(\widetilde{X}^{s}\right), f_{\theta i}\left(\widetilde{X}^{t_{i}}\right)\right)+
\\ \max _{i} d\left(f_{\theta i}\left(\widetilde{X}^{t_{i-1}}\right), f_{\theta i}\left(\widetilde{X}^{t_{i}}\right)\right)
\end{aligned}
&&
\end{flalign}
Therefore, in the outer loop, the parameters of the feature extractor are updated according to the following formula:
\begin{flalign}\label{outer}
&&
\begin{aligned}
(\varphi,\mu_{e},\mu_{d})_{i+1} \leftarrow(\varphi,\mu_{e},\mu_{d})_{i}  -{\alpha_{out} }\nabla_{\varphi,\mu_e,\mu_d} L_{out}
\end{aligned}
&&
\end{flalign}
\end{small}
where $\varphi$ is the parameter of the feature extractor, the $\mu_e$ and $\mu_d$ are the parameters of the auto-encoder $A$, $i$ is the number of iteration rounds for the outer loop and $\alpha_{out} $ is the learning rate of the outer loop.
The complete meta-training process is given in Algorithm \ref{alg:1}.

\begin{algorithm}
	\caption{Meta-training of PCAda}
    \begin{algorithmic} 
	\label{alg:1}
		\REQUIRE the source domain $D_s$ and the evolving target domain $D_t$.
  
		\ENSURE prototype $\mathbf{c}_{k}$, encoder $A_e$, classifier $f_\theta$ and feature extractor  $\phi$.
  
		\STATE Initialize encoder $A_e$, decoder $A_d$, feature extractor $\phi$ and classifier $f_\theta$.
  
		\STATE Pre-training model by using $D_s=\{X^s,Y^s\}$.
  
        \STATE Initialize prototype $\mathbf{c}_{k}$ by Equation~\ref{prototype}.
        
		\FOR{t=0 to MaxOuter}
  
		\STATE Sample target trajectories $D^{spt}_t= \{\hat{X}^{t1},\hat{X}^{t2},…,\hat{X}^{tm}\}$ and $D^{qry}_t  = \{\widetilde{X}^{t1},\widetilde{X}^{t2},…,\widetilde{X}^{tm}\}$ from $D_t$.
		\FOR{i=0 to MaxInner}
		\STATE Update prototype $\mathbf{c}_{k}$ with $D^{spt}_t= \{\hat{X}^{t1},\hat{X}^{t2},…,\hat{X}^{tm}\}$  based on Equation~\ref{prototype_dis}, \ref{pseudo-label}, \ref{update}\STATE Update classifier $f_\theta$ based on Equation~\ref{inner}
        \ENDFOR
        \STATE Use $D^{qry}_t  = \{\widetilde{X}^{t1}, \widetilde{X}^{t2},…, \widetilde{X}^{tm}\}$, encoder $A_e$ and decoder $A_d$ to generate mask $A_i$
        \STATE Update feature extractor $\phi$  , encoder $A_e$ and decoder $A_d$ with mask $A_i$ based on Equation~\ref{mask}, \ref{outer}
		\ENDFOR
  \end{algorithmic} 
\end{algorithm}

\section{Experiment}

In this section, we use evolving domain datasets in different scenarios to evaluate our method, and analyze whether(and why) the model can effectively capture the knowledge transfer and mitigate forgetting.

\subsection{Datasets}
We will modify the existing datasets to emulate an evolving domain shift situation. More details are shown in the appendix.

\textbf{Rotated MNIST~\cite{lecun1998mnist}}: We prepared the rotated MNIST datasets, following the previous work~\cite{liu2020learning}, the rotation of the target domain is continuous  from $0^{\circ}$ to $180^{\circ}$. 
Images without rotation belong to the labeled source domain. For meta-training, we select data between 0 and 60 degrees, $3^{\circ}$ as intervals, and 20 evolving target domains in total. 10 target domains are randomly selected among them and sorted according to the rotation degrees as $D_t=\{X^{t_1},X^{t_2},...,X^{t_{10}}\}$, $t_i \in [0,60]^{\circ}$. 
For meta-testing, we test the model’s performance online with trajectory $D_t=\{X^{120^{\circ}},X^{126^{\circ}},...,X^{174^{\circ}}\}$.

\textbf{Caltran~\cite{hoffman2014continuous}}: This is a surveillance dataset of images captured by a fixed traffic camera observing an intersection for scene classification. 
The dataset includes images captured at an interval of 3 minutes over 2 weeks. Different from ~\cite{liu2020learning}, our setup is as follows: 
In the first 7.5 hours, images are set as the labeled source domain, and the continually evolving target domain contains images in the following two weeks with the same time interval of 7.5 hours. 
For meta-training, we selected the data of the first week and divided it into 18 evolving domains, 9 target domains are randomly selected among them and sorted according to time as $D_t=\{X^{t_1}, X^{t_2},..., X^{t_{9}}\}$. For meta-testing, we use the data of the second week.
\begin{table*}[!htp]
\centering
\caption{\textbf{Classification Accuracy (\%) on rotated MNIST dataset}} \label{tab:t1} 
\scalebox{0.8}
{
\begin{tabular}{cccccccccccc}
\toprule
Method&$120^{\circ}$&$126^{\circ}$&$132^{\circ}$&$138^{\circ}$&$144^{\circ}$&$150^{\circ}$&$156^{\circ}$&$162^{\circ}$&$168^{\circ}$&$174^{\circ}$& Average \\
\midrule
Source Only&17.60&19.29&22.50&24.14&26.49&29.48&31.06&32.26&33.73&33.25&26.98 \\
DANN\cite{ganin2015unsupervised}&18.92&21.65&24.32&27.63&29.76&32.01&33.92&36.23&36.68&36.93&29.81\\
JAN\cite{long2017deep}&20.37&22.53&25.15&27.53&29.89&31.40&32.99&33.85&35.77&37.13&29.66 \\
GDAMF\cite{sagawa2022cost}&24.39&26.48&27.81&\textbf{35.40}&34.96&36.56&42.43&41.60&39.87&44.73&35.42 \\
EAML\cite{liu2020learning}&24.69&27.48&30.16&32.79&34.88&37.35&39.25&40.96&42.45&42.27&35.23 \\
PCAda&\textbf{25.25}&\textbf{28.83}&\textbf{32.26}&35.18&\textbf{38.4}&\textbf{40.88}&\textbf{43.01}&\textbf{44.41}&\textbf{45.26}&\textbf{45.91}&\textbf{37.95} \\
\bottomrule
\end{tabular}
}
\end{table*}

\begin{table*}[!htp]
\centering
\caption{\textbf{Classification Accuracy (\%) on Caltran and Portraits}}\label{tab:t2}
\scalebox{1.0}{
\begin{tabular}{ccccc|cccc}
\toprule
\textbf{Dataset}&\multicolumn{4}{c}{\textbf{Caltran}}&\multicolumn{4}{c}{\textbf{Portraits}}\\
%\midrule
\hline
Method&Source Only&JAN&EAML&PCAda&Source Only&JAN&EAML&PCAda \\
% \midrule
\hline
Average   &  42.56 &  81.63 &  84.04 &  \textbf{89.67} &    48.27 &  67.88 &  76.88 &    \textbf{80.49} \\
% \bottomrule
\hline
\end{tabular}}
\end{table*}

\begin{figure}{
\includegraphics[width=9cm]{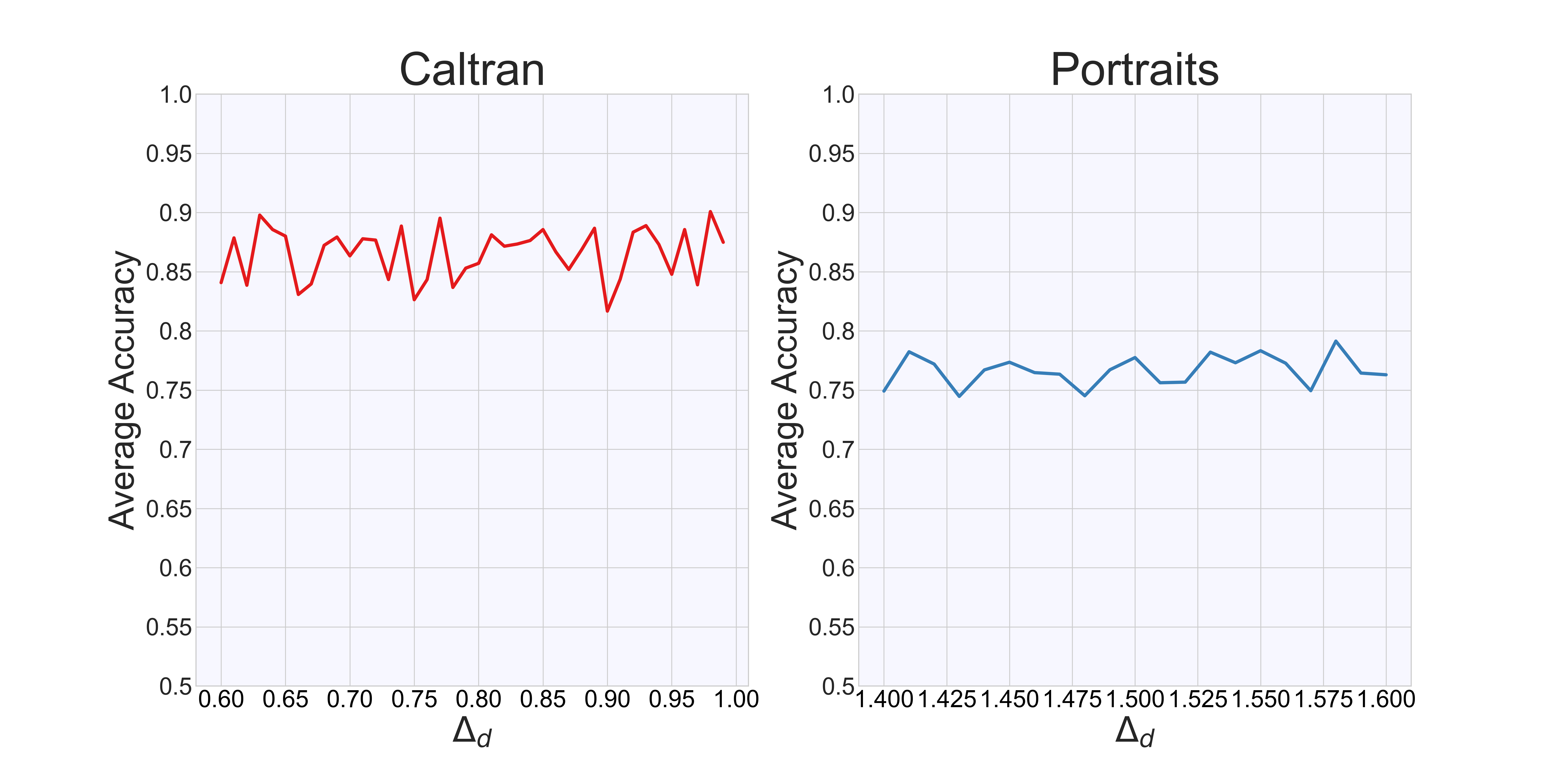}
\caption{\textbf{Hyperparameter Sensitivity.}}
\label{hyper}}
\end{figure}

\textbf{Portraits~\cite{ginosar2015century}}
Portraits dataset is a real-world gender classification dataset of a collection of American high school seniors from the year 1905 to 2013. 
Our task's goal is to predict the gender from the images. Since the portrait styles change over the years, these images can be sorted in ascending order by years and split into 36 continually evolving target domains. 
Each domain contains 1000 images and the first domain is set as the labeled source domain. 
In meta-training, we sampled 9 of the first 18 domains and ranked them in time order as $D_t=\{X^{t_1},X^{t_2},...,X^{t_{9}}\}$ and the batch size is 16. 
For meta-testing, we test the last unused 18 domains.

\subsection{Experimental Settings}
In the case of the rotating MNIST, we use LeNet as the backbone, and the feature extractor network $\phi$ was composed of two convolutional layers. The encoder $A_e$ and decoder $A_d$ are both one-layer MLP. The classifier $f_{\theta}$ is a two-layer fully-connected network with ReLU activations. In the case of Caltrans and Portraits, we use ResNet18 as the backbone. Others network in the same as the case of the rotating MNIST. We used SGD with 0.9 momentum as the optimizer. For the inner loop, the learning rate $\eta_{f}$ was set to 0.01, and for the outer loop, the learning rate and weight decay were both set to 0.001.
$\Delta_d$ is set to 0.8 for Caltran and 1.5 for Portraits. In Equation \ref{tuihuo}, $T_1$ is set to 20 and $T_2$ is set to 40.

\begin{figure*}[!htp]
	\centering
	\scalebox{0.95}{
{\includegraphics[width=.245\textwidth]{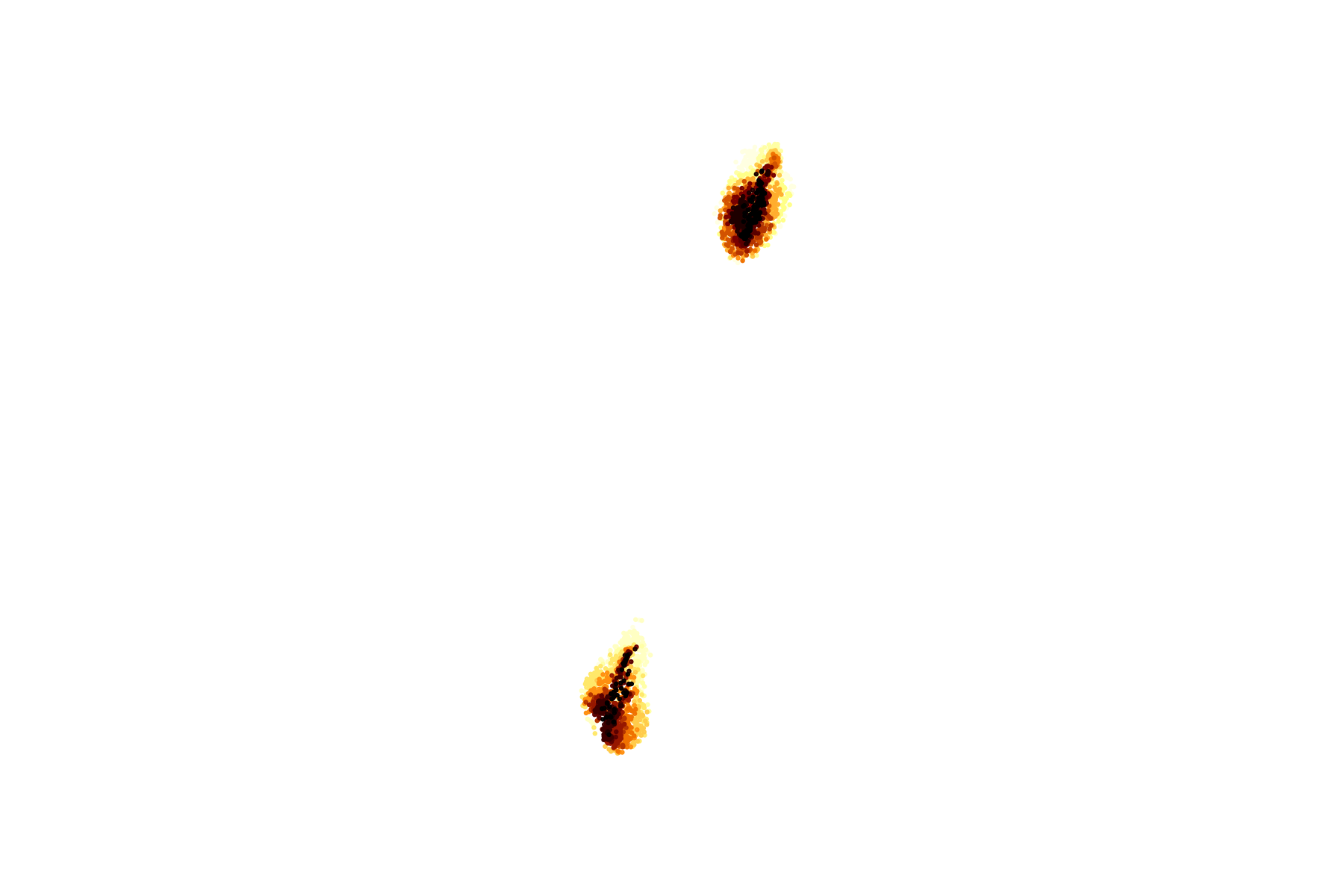}}
{\includegraphics[width=.245\textwidth]{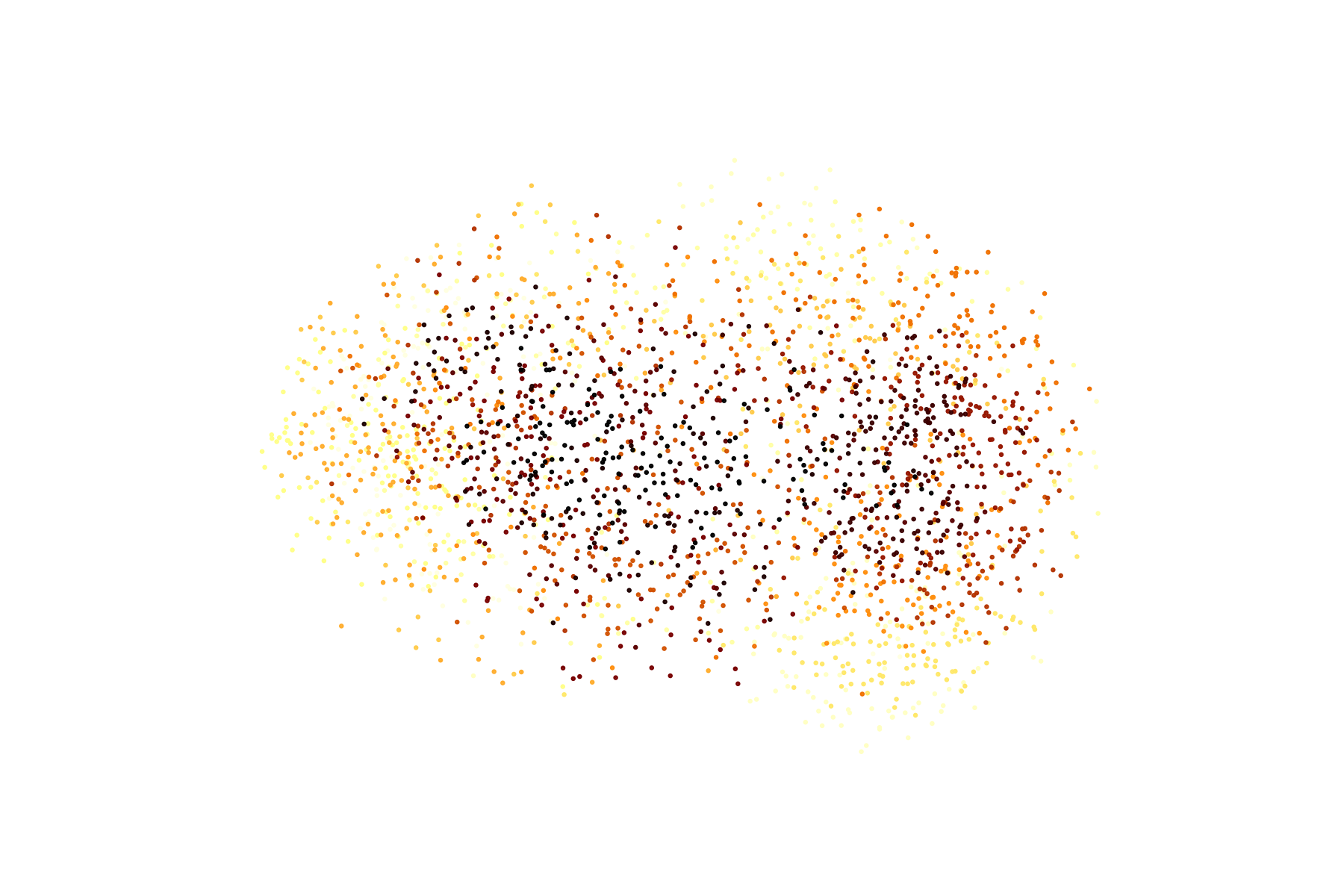}}
{\includegraphics[width=.245\textwidth]{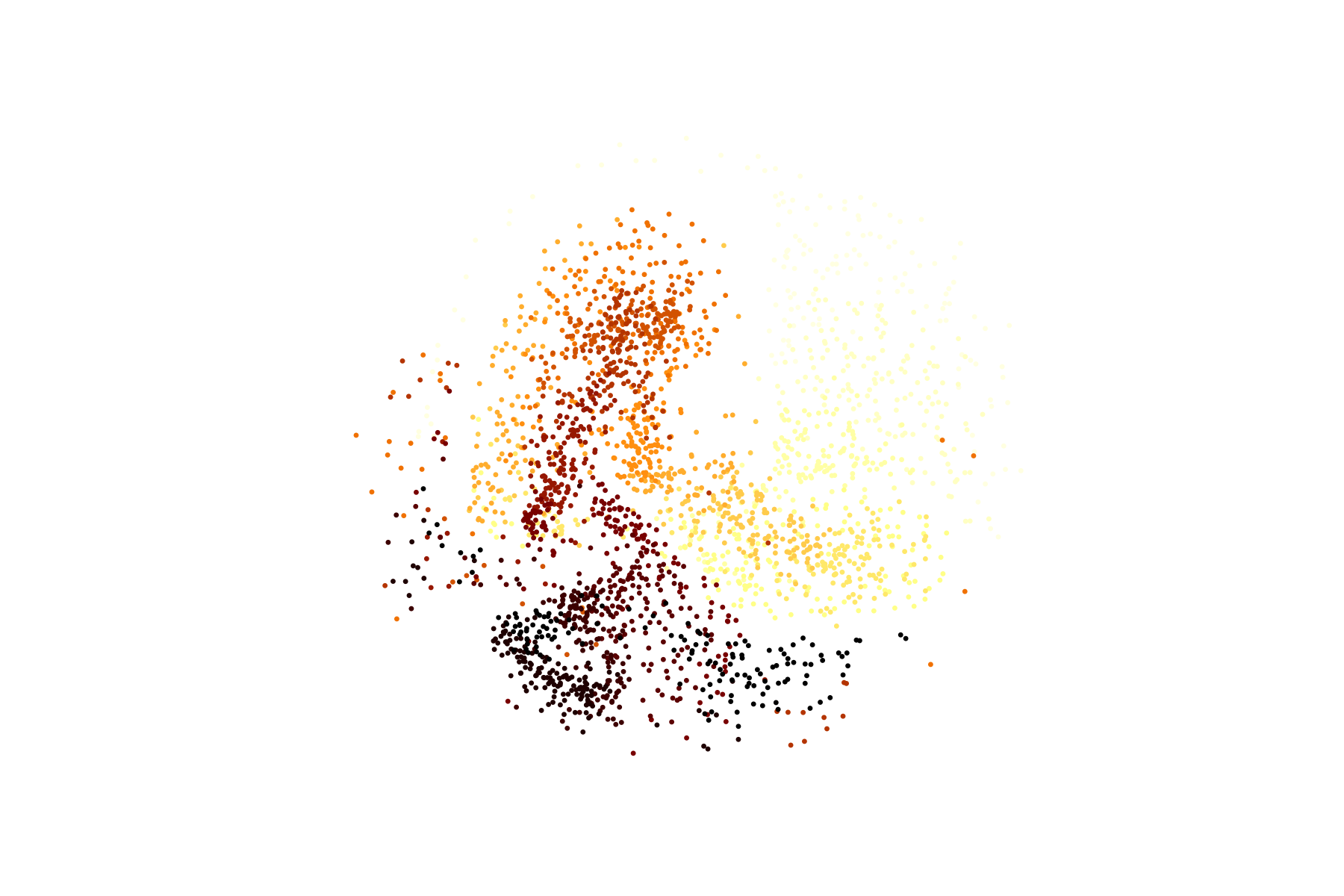}}
{\includegraphics[width=.245\textwidth]{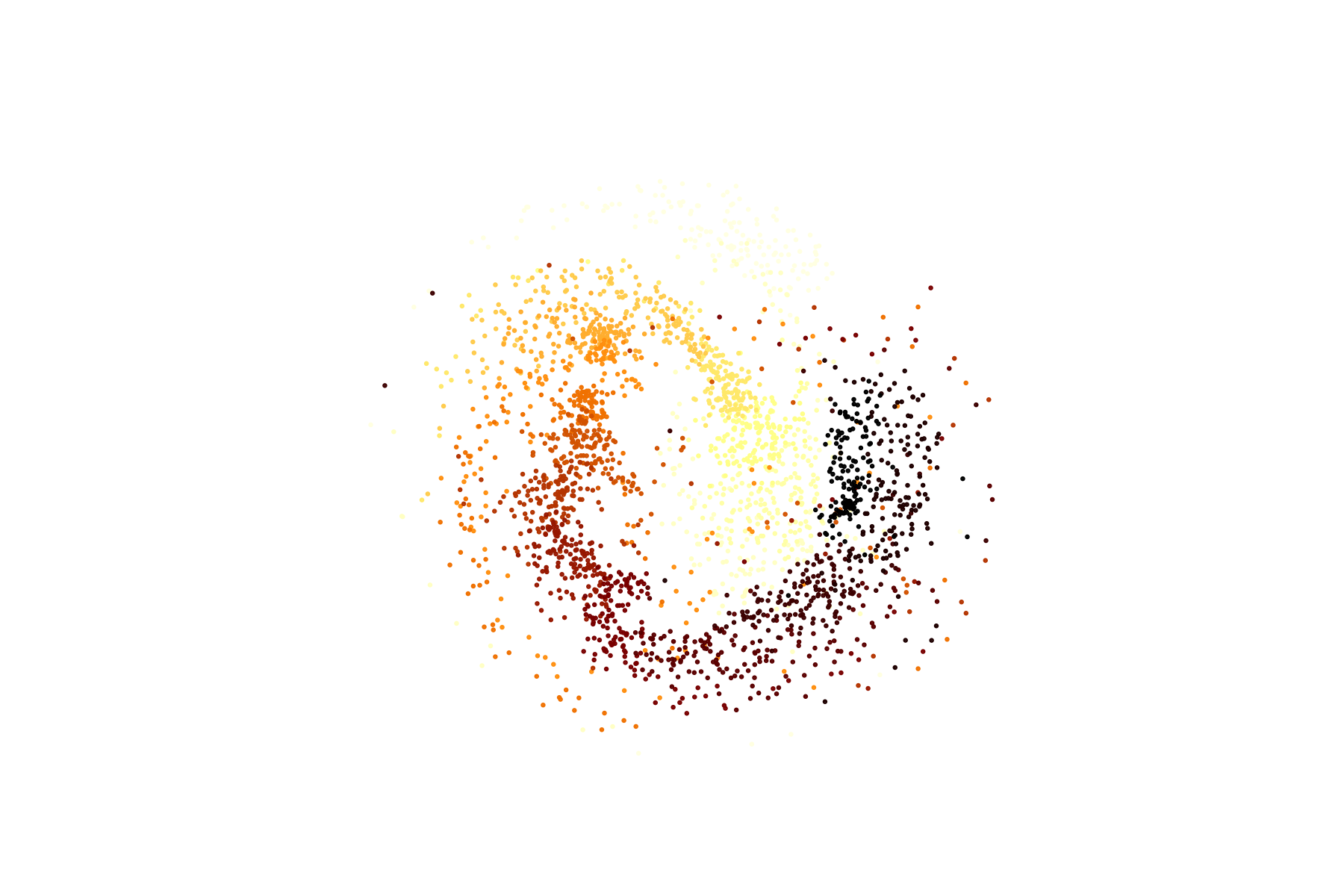}}}
\caption{\textbf{UMAP on Caltran.} As the color changes, the target domain continues to evolve. The model is trained only in the source domain and JAN cannot capture the evolution of the target domain. EAML does not learn very well about the special properties of data in evolution, such as periodicity. From left to right, it shows the results from source, jan, eaml, and ours. As shown that our method is capable of continually learning evolving knowledge.}
\label{umap}
\end{figure*}

\subsection{Analysis and Results}
In this section, we will present the results on all datasets and analyze the details. 
%The complete experimental results are shown in the Appendix.
% {\normalfont\normalsize\CJKfamily{hei}}}
\subsubsection{Baselines}
We want to show that PCAda can make more accurate predictions with better performance to prevent forgetting than existing baseline methods. We include DANN~\cite{ganin2016domain},  JAN~\cite{long2017deep}, GDAMF~\cite{sagawa2022cost},and EAML~\cite{liu2020learning} in the comparative study because the implementations are publicly available. The experimental setup for each baseline method is shown below.\\
\textbf{Source only}: Train only on the source dataset. And test on all evolving target domains. \\
\textbf{DANN~\cite{ganin2015unsupervised} and JAN~\cite{long2017deep}}: Two classic domain adaptation methods.
We just train the model with the source dataset and all target datasets sequentially.
Keep the same setup as our method.   \\
%值得注意的是，此处部分目标域是可以有少数标签的
\textbf{GDAMF~\cite{sagawa2022cost}}: The state-of-the-art gradual domain adaptation method. 
A few data in some target domains are labeled.
So this method can benefit from this setting.
Other setup are the same as our method.\\
\textbf{EAML~\cite{liu2020learning}}: The the state-of-the-art EDA method. 
We use this method in our experimental setup. 

\subsubsection{Results on Evolving Domain}
Table \ref{tab:t1} shows the results on the Rotated MNIST and Table \ref{tab:t2} shows the results on the Caltran and Portraits dataset. 
We compare the prediction accuracy of each target domain and show the results in the Appendix.
As illustrated in Tables \ref{tab:t1} and \ref{tab:t2}, our method achieves the new state of the art on all evolving target domains and benchmark datasets, justifying its efficacy on EDA problems and its generalization to different datasets.

\subsubsection{Ablation Study} We investigate the Adaptive Prototype Mechanism (APM) and Sparse Attention Mechanism (SAM) by removing the corresponding component from the overall objective. By observing the experimental results, we found that APM and SAM performs better than basic ways. 

\noindent\textbf{The effect of APM and SAM}. To further show the effect of our method to prevent forgetting, following previous works, we use two key metrics in Continual learning: Average Accuracy (ACC) and Backward Transfer (BWT)~\cite{lopez2017gradient}.
\begin{small}
\begin{flalign} \label{eq_acc}
&&
ACC  = \frac{1}{T} \sum_{i=1}^{T}R_{T,i}
&&
\end{flalign}\begin{flalign} \label{eq_bwt}
&&
BWT = \frac{1}{T-1} \sum_{i=1}^{T-1}R_{T,i}-R_{i,i}
&&
\end{flalign}
\end{small}
where $R$ is the test classification accuracy, $T$ is the number of total domains, the same as the number of tasks and the $i$-th task means that we only use the first $I$ domains for meta-testing.
The Average Accuracy measures the average accuracy of each domain after the end of the meta-testing process. 
BWT indicates the effect of how learning a domain $t$ influences the performance of other domains $k<t$.
Specifically, a low ACC score and a negative BWT score are referred to as catastrophic forgetting, and a high ACC score and a positive BWT score imply the model is improving its overall learning when it is trained using a new domain.
%前i个域

\noindent\textbf{Hyperparameter Sensitivity}. Average accracy on Caltran and Portraits with respect to different values of hyperparameter $\Delta_{d}$ is shown in Figure \ref{hyper}. Results confirm that our method is not sensitive to hyperparameter.

%\noindent\textbf{Batch-sum VS. without it}. We investigate how batch-sum operation impacts the accuracy over each target domain. Figure~\ref{fig:role} show that using batch sum operation attains higher accuracy for all domains and our method has lower standard deviations. The possible reason is generating the same mask will make the method update more stable for the data in a batch.
%\begin{figure*}[!htp]
%	\centering
%	\scalebox{0.95}{
%{\includegraphics[width=9cm]{Figure/BN2(caltran).png}}
%{\includegraphics[width=9cm]{Figure/BN2(face).png}}}
%\caption{From left to right, it shows \textbf{The Role of the Batch sum} on Caltran and Portraits datasets. Using batch sum operations can generate uniform and similar masks over the entire batch of data and thus make the model update method more stable in evolving domains.}
%\label{fig:role}
%\end{figure*}

\begin{figure}[!htp]{
\includegraphics[width=8cm]{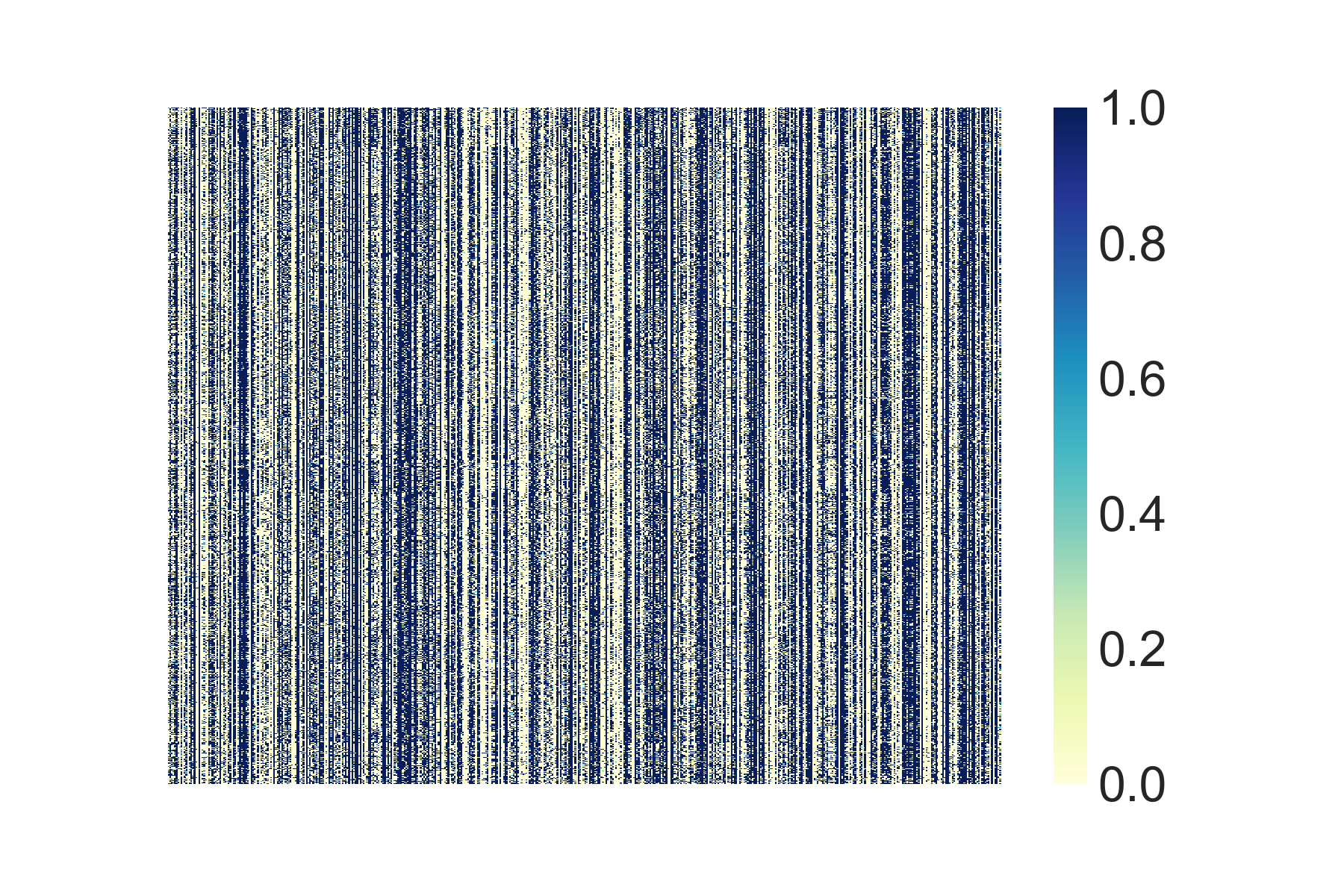}
\caption{\textbf{The Visualization of SAM.}}
\label{heatmapsam}}
\end{figure}

We compute the ACC and BWT of these methods on Caltran and Portraits.
We conduct 20 trials and Mean and standard deviations are reported.
The best results in each column are highlighted in  \textbf{bold} font.
Results are provided in Table~\ref{tab:t3}. It can be seen that the two methods proposed in this paper have better improvement in ACC and BWT compared with the baseline, which proves the ability of our method to reduce forgetting. Even though the BWT decreased slightly in the process of combining the two methods, the ACC was greatly improved, which reflected a balance between the model's domain adaptation and overcoming the forgetting. We use UMAP~\cite{mcinnes2018umap} to map features learned by different methods on Caltran to 2D space in Figure \ref{umap}. The color changes from light to dark as the data evolves from early target domains to old target domains.  Training only on the source domain didn't show any signs of domain evolution. JAN and EAML can capture smooth evolving target domain shifts to a different extent. But these methods are not always accurate, a lot of data from early evolving target domains mix with the data from later domains. The evolution of the domain in PCAda is relatively more orderly. Then we take the sparse attention corresponding to each data of a domain in the portraits for heatmap visualization in Figure \ref{heatmapsam}. We can see obvious stripes in the map, which shows that our method has obtained an appropriate sparse attention representation for the data of the same domain.

\section{Conclusion}
In this paper, to solve the evolving domain adaptation (EDA), we propose a meta-learning framework model, which consists of two parts: First, for the EDA problem in which the model needs to quickly adapt to the evolution of target data, we design a progressive prototype update mechanism enables the classifier to quickly adapt to the evolving target domain. Second, to help the model not forget historical information, we propose a conservative sparse attention mechanism. Focus on updating the key feature channels in the current domain, and reduce the interference of channels that play a decisive role in historical domain features. Experiments on three public benchmarks including Rotated MNIST, Caltran, Portrait demonstrate our model outperforms existing EDA models in evolving domains. All visualization results show that our method achieved the best performance, such as in Figure \ref{heatmapsam}, the proposed method has obtained an appropriate sparse attention representation for the data of the same domain. We hope our work can help future research and further advance the real-world application of domain adaptation.

\appendix
% {\small
% \bibliographystyle{aaai24}
% \bibliography{aaai24}
% }

\bibliography{aaai24}

\end{document}